# A Hybrid Anytime Algorithm for the Construction of Causal Models From Sparse Data.

**Denver Dash[†]**
Department of Physics and Astronomy
and Decision Systems Laboratory
University of Pittsburgh
Pittsburgh, PA 15260
ddash@sis.pitt.edu

**Marek J. Druzdzel**
Decision Systems Laboratory
School of Information Sciences
University of Pittsburgh
Pittsburgh, PA 15260
marek@sis.pitt.edu

[†] Mr. Denver Dash is a full time doctoral student in the Decision Systems Laboratory at the University of Pittsburgh.

## Abstract

We present a hybrid constraint-based/Bayesian algorithm for learning causal networks in the presence of sparse data. The algorithm searches the space of equivalence classes of models (essential graphs) using a heuristic based on conventional constraint-based techniques. Each essential graph is then converted into a directed acyclic graph and scored using a Bayesian scoring metric. Two variants of the algorithm are developed and tested using data from randomly generated networks of sizes from 15 to 45 nodes with data sizes ranging from 250 to 2000 records. Both variations are compared to, and found to consistently outperform two variations of greedy search with restarts.

## 1 Introduction

Methods for learning probabilistic graphical models can be partitioned into at least two general classes of methods: constraint-based search and Bayesian methods. The constraint-based approaches [Spirtes et al., 1993, Verma and Pearl, 1991] search the data for conditional independence relations from which it is in principle possible to deduce the Markov equivalence class of the underlying causal graph. Two notable constraint-based algorithms are the PC algorithm which assumes that no hidden variables are present and the FCI algorithm which is capable of learning something about the causal relationships even assuming there are latent variables present in the data [Spirtes et al., 1993]. Bayesian methods [Cooper and Herskovits, 1992] utilize a search-and-score procedure to search the space of dags, and use the posterior density as a scoring function. There are many variations on Bayesian methods (see [Heckerman, 1998] for a comprehensive review of the literature); however, most research has focused on the application of greedy heuristics, combined with techniques to avoid local maxima in the posterior density (e.g., greedy search with random restarts or best-first searches).

Both constraint-based and Bayesian approaches have advantages and disadvantages. Constraint-based approaches are relatively quick and possess the ability to deal with latent variables. However, constraint-based approaches rely on an arbitrary significance level to decide independencies, and they can be unstable in the sense that an error early on in the search can have a cascading effect that causes a drastically different graph to result. Bayesian methods can be applied even with very little data where conditional independence tests are likely to break down. Both approaches have the ability to incorporate background knowledge in the form of temporal ordering, or forbidden or forced arcs, but Bayesian approaches have the added advantage of being able to flexibly incorporate users' background knowledge in the form of prior probabilities over the structures and over the parameters of the network. Also, Bayesian approaches are capable of dealing with incomplete records in the database. The most serious drawback to the Bayesian approaches is the fact that they are relatively slow.

Typically Bayesian search procedures operate on the space of directed acyclic graphs (dags). However, recently researchers have investigated performing greedy Bayesian searches on the space of equivalence classes of dags [Spirtes et al., 1995, Madigan et al., 1995, Chickering, 1996]. The graphical objects representing equivalence classes have been called by several names ("patterns", "completed pdag representations", "maximally oriented graphs", and "essential graphs"). We will use the term "essential graph" because we feel it is both descriptive and concise (but we acknowledge that the term "pattern" is more prevalent). An essential graph is a special case of a chain graph, possessing both directed and non-directed arcs, but no directed cycles. In order to specify an equivalence class it is necessary and sufficient to specify both a set of undirected adjacencies and a set of v-structures (a.k.a. "non-shielded colliders", a structure such as $X \to Y \leftarrow Z$ such that $X$ is not adjacent to $Z$) possessed by the dag [Chickering, 1995]. An essential graph therefore possesses undirected adjacencies when two nodes are adjacent, and it may possess directed adjacencies if a triple of nodes possesses a v-structure or if an arc is required to be directed due to other v-structures [Anderson et al., 1995]. The space of essential graphs is smaller than the space of dags; therefore it is hoped that performing a search directly within this space might be beneficial; however, the Bayesian metric must be applied to a dag, therefore these procedures incur the additional cost required to convert back and forth between essential-graph-space and dag-space. Results from the above work have shown to be promising, however.

Researchers have also developed two-stage hybrid algorithms, where the first stage performs a constraint-based search and uses the resulting graph as input into a second-stage Bayesian search. In particular, [Singh and Valtorta, 1993] use the PC algorithm to generate an absolute temporal ordering on the nodes for use with the K2 algorithm [Cooper and Herskovits, 1992], which requires such an ordering on the input. [Spirtes et al., 1995] use the PC algorithm to generate a good starting graph for use in their greedy search over the space of essential graphs.

We describe in this paper two new variants of a hybrid constraint-based/Bayesian algorithm that is effective for sparse data. The algorithms use conventional constraint-based methods to search the space of essential graphs, convert these to dag-space, and then proceed in different ways to apply Bayesian techniques. In Section 2 we review the PC algorithm and the Bayesian scoring metric. In Section 3 we describe the new algorithms in detail. Finally, in Section 4 we describe the design and results of some experimental tests that were used to analyze the algorithms, including some comparisons to a conventional Bayesian greedy search algorithm and to a hybrid constraint-based/Bayesian algorithm.

## 2 The PC Algorithm and the Bayesian Criterion

The PC algorithm is a constraint-based algorithm for finding the essential graph corresponding to the causal graph which generated the data. This algorithm assumes the *Causal Markov Condition* and the *Causal Faithfulness Condition* as well as valid statistical decisions. We sketch this algorithm below. For a detailed description the reader is referred to [Spirtes et al., 1993].

The procedure takes as input a database over a set of variables $V$, a test of conditional independence $I(x, y|S)$, and a significance level $0 < \alpha < 1$. Implicitly the algorithm also takes an ordering $order(V)$ over the nodes which specifies in which order to check for independencies.

**Procedure 1 (PC Algorithm (sketch))**

**Input:** A database $D$ over a set of variables $V$, a test of conditional independence: $I(x, y|S)$, a significance level: $0 < \alpha < 1$, and an ordering $order(V)$ over $V$.

**Output:** An essential graph over $V$.

1. Construct the complete undirected graph over $V$.

2. For all adjacent nodes $x$ and $y$, try to separate nodes by checking first for lower-order, then for progressively higher-order conditional independencies between $x$ and $y$. Check a conditional independence relation $I(x, y|S)$ if and only if all variables in $S$ are adjacent to either $x$ or $y$. If a conditional independence relation is discovered between $x$ and $y$, then remove the edge between $x$ and $y$, thus decreasing the number of possible sets, $S$. Conditional independencies should be checked in the order specified by $order(V)$.

3. For each triple of nodes $(x, y, z)$ such that $x$ is adjacent to $y$ and $y$ is adjacent to $z$ but $x$ is not adjacent to $z$, Orient $x$—$y$—$z$ as $x \to y \leftarrow z$ if and only if $y$ was not in the set $S$ that separated $x$ and $z$ in step 2.

4. Repeat, until no more edges can be directed:
   (a) Direct all arcs necessary to avoid new v-structures.
   (b) Direct all arcs necessary to avoid cycles.

The specification of $order(V)$ is not typically emphasized in the statement of the PC algorithm; however, it will be important for us later, so we explicate it here. It should be noted that $order(V)$ does not specify a *temporal* ordering over the variables. It merely is an arbitrary parameter that specifies in what order to check for independencies. In the infinite sample limit, the output of the algorithm would be independent of these different orderings.

Steps 1 and 4 are performed in $O(n^2)$ time. Steps 2 and 3, however are exponential in $n$, in the worst case.

The worst case occurs when the generating graph is very dense; in particular, the computational time complexity of PC is bounded by $n^2(n-1)^{k-1}/(k-1)!$, where $k$ is the maximum degree of any node in the graph. This expression provides a loose upper bound on the worst-case complexity; the average-case complexity depends on the details of the structure of the underlying causal graph.

The Bayesian metric that we use was first derived by [Cooper and Herskovits, 1992]. This score is the joint density $P(D,S) = P(D|S)P(S)$ which is proportional to the posterior density $P(D,S) \propto P(S|D)$. Under certain assumptions about the form of the priors over the parameters of the model and under the assumptions of parameter independence (see [Heckerman, 1998] for details of these assumptions and other issues) the logarithm of $P(D,S)$ can be calculated efficiently and in closed form.

## 3 Description of the Algorithms

As stated previously, the PC algorithm is unstable. In both step 2 and in step 3 an error made early on can have cascading effects in the output. This effect is especially apparent for sparse data sets with many variables. [Spirtes et al., 1993] note this instability and remark that in practice step 2 is much more stable than step 3. Because of these instabilities, the output of the PC algorithm is sensitive to the order in which conditional independencies are checked. Of course the output is also sensitive to the value of the significance level, $\alpha$ specified by the input. The algorithms described here capitalize on these sensitivities in the PC algorithm to perform a search through the space of essential graphs; namely, by varying these arbitrary parameters it is possible to generate many different essential graphs with the same set of data.

An example of this effect is shown in Figure 1. The PC algorithm was run on 4 databases ranging from 250 to 2000 records containing 15 variables each. Each database was generated by a different randomly generated graph (see Section 4 for details about how the data-generating graphs were created). PC was run on each of these four data sets 1300 times. After each termination of PC, the $order(V)$ was changed randomly and a new $\alpha$ was drawn from a uniform distribution in the range from 0.005 to 0.2, and PC was run again. This procedure was repeated 1300 times per data set. This figure depicts the number of distinct graphs generated out of 1300 attempts. For example, in the database with 1000 records, the PC algorithm generated almost 450 distinct essential graphs out of 1300 attempts. Clearly, this procedure is effective for searching the space of essential graphs. However, Fig-ure 1 indicates that the ability to search the space begins to decrease as the number of records increases. This is because the statistical decisions become more robust with increasing data and make fewer errors. Conversely, since the space of essential graphs grows rapidly as the number of variables grows, the number of distinct graphs generated should also sharply increase with the number of variables.

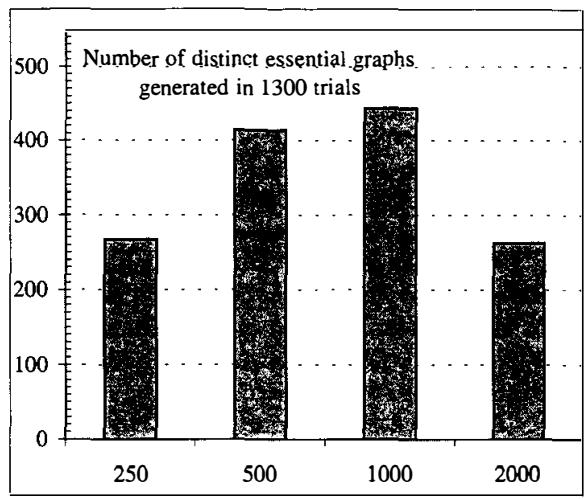

Figure 1: After 1300 runs of the PC algorithm, several distinct essential graphs were generated by varying the significance level and $Order(V)$. The databases for 15 variables had numbers of records equal to 250, 500, 1000, and 2000, respectively.

The first procedure we describe performs a search of essential graph space using the PC algorithm as a heuristic, then randomly converts each essential graph to a dag and scores the dag using the Bayesian metric. It is an "anytime" algorithm in the sense that the longer the algorithm runs the better its output will be. We will call this algorithm EGS for "essential graph search".

**Procedure 2 (EGS)**

**Input:** A database $D$ over a set of variables $V$, a distribution $P(\alpha)$ over the significance level $\alpha$, a test of independence $I(x,y|S)$, an initial configuration for $order(V)$, and an integer $n$.

**Output:** An essential graph.

**Repeat:**

1. Draw an $\alpha$ from $P(\alpha)$.
2. Call $PerformPC(\alpha, order(V), D, I)$ to generate an essential graph $G'$.

3. Randomly convert $G'$ to a dag $S'$.
4. Calculate $P(D, S')$ and record the structure with the maximum value.
5. Randomly generate a different configuration of $order(V)$.

**Until:** $n$ structures are generated without updating the maximum value of $P(D, S')$.

Many variations of this algorithm are possible. Since the metric being used is not invariant over dags in the same equivalence class, it may be more reasonable to generate and score all possible dags from a given essential graph, rather than randomly picking a single one. Another possibility is to use a scoring function that is invariant under equivalence class transformations. [Heckerman et al., 1995] suggests one possibility for an invariant metric. We conjectured that our results would not be too sensitive to these factors.

The second algorithm is a variation on EGS which inserts in place of step 4 an entire greedy search procedure over the space of dags. It is thus a form of greedy search with restarts that uses EGS to generate the graphs to use at each new restart interval. We call this algorithm EGS/GS, for "EGS/Greedy Search" variation".

Again, many variations are possible. In our experiments, all greedy search procedures used the same basic procedure: Some initial graph is first selected as a starting point. From there three search operators are tried in succession: $Add(x, y)$ adds an arc between $x$ and $y$, $Del(x, y)$ deletes an arc between $x$ and $y$, and $Rev(x, y)$ reverses an arc between $x$ and $y$. The results of these operations are then calculated for each pair of nodes $(x, y)$, and the operation that results in the largest increase in score was chosen. Another possibility is to use the K2 algorithm as a basis for the greedy search.

These algorithms assume for simplicity that no hidden variables are present. Unfortunately, it is possible that the PC algorithm will detect some hidden variables in the course of its search through essential graph space. This can happen when directing arcs in step 3 of the PC algorithm. In particular, if two triples of variables exist: $A$—$B$—$C$ and $B$—$C$—$D$, and PC simultaneously detects two v-structures $A \to B \leftarrow C$ and $B \to C \leftarrow D$, then PC will create a structure $B \leftrightarrow C$, which can be interpreted as a hidden common parent of $B$ and $C$. Although this feature is useful in general, we do not deal with it here. Therefore, step 3 of the PC algorithm was modified so that it never added a double-headed edge to the graph. This was accomplished by checking triples in an order specified by $order(V)$ and simply prohibiting the second head of any double-headed arcs whenever they appear.

It is also possible, but much more rare, for step 3 of PC to create a directed cycle in the graph. This can occur when a "pinwheel" structure is found as shown in Figure 2. Again, for convenience we reject graphs containing these structures.

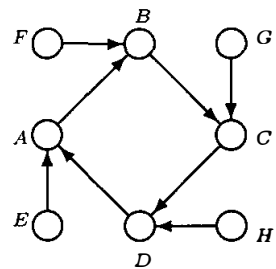

Figure 2: A "pinwheel" structure than can produce cycles when orienting arcs in Step 3 of the PC algorithm.

One might question the necessity or the validity of manipulating the significance level parameter in an arbitrary way, as we are suggesting here. After all, well established rules have been developed in the statistics community over the decades to assist us in choosing meaningful significance levels for hypothesis testing. Also, [Spirtes et al., 1993] have developed guidelines based on extensive experimental tests for which significance levels to choose in the course of constraint-based causal discovery.

We propose both theoretical and experimental justification for searching over the significance level. On theoretical grounds, we argue that the process of searching for a causal graph using tests for significance is a considerably different procedure than the standard hypothesis testing that has been studied in statistics for decades. In a typical hypothesis test, say a student t-test, the significance level is used to judge the probability of a single hypothesis. By contrast, the number of hypotheses tested in a typical constraint-based search is typically a polynomial function of the number of variables in the model. Therefore, it is much more likely that one of these hypothesis tests will fail by chance and an arc will be erroneously added or deleted from the model. This fact, coupled with the fact that constraint-based searches tend to be unstable in errors made early-on in the search, means that constraint-based searches are much more sensitive to the significance level than traditional hypothesis tests.

To empirically demonstrate the importance of varying the significance level during the search, Figure 3 shows the number of graphs of maximal quality (as measured by the Bayesian metric) found as a function of significance level for the EGS algorithm performed on 50 databases of 500 records generated by a set of 50 graphs having 15 nodes and mean number of arcs equal to 22. It can be seen that although clearly the significance of 0.01 was the most important, over 70% of the maximal quality graphs generated were generated at other significance levels over the course of these EGS runs.

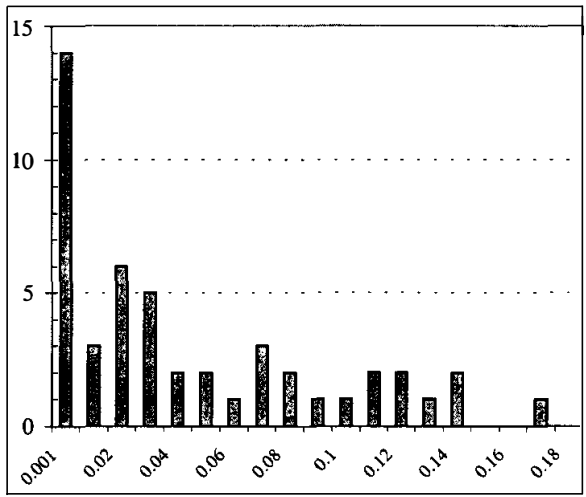

Figure 3: Histogram of the significance levels that produce the essential graphs with maximum score with the EGS convergence parameter set to $n = 500$. The generating graph consisted of 15 nodes, with $\overline{M}_{arcs} = 22$ and was used to generate 500 records.

## 4 Experiments

We performed three experiments comparing EGS and EGS/GS to two variations of greedy search with restarts. Both variations of greedy search operate on the space of dags and explore the space using the operations described in Section 3. Both variations score the resulting structure using the metric $P(D, S)$ and at each step pick the operation that maximizes this quantity. All greedy search procedures reported in this paper also take advantage of the separability of the $P(D, S)$ metric so that only portions of the metric relevant to the nodes that have changed need to be recalculated at each step (See [Heckerman, 1998] for details of separable criteria).

The first variant of greedy search, which we call "GS" begins the search with a random graph and at each restart interval generates a new random graph. The second variant "GS/1" uses the PC algorithm to initiate the search, then at each subsequent restart a new dag is generated randomly.

In all of the experiments described below, random data was generated using randomly constructed graphs and linear standard-normal ($\mu = 0$, $\sigma = 1$) models with random parameters drawn from a uniform distribution between 0.1 and 0.9. That is, if a variable $Y$ had a single parent $X$ in the model, then the value for $Y$ was generated to be $Y = \beta X + \gamma$, where $0.1 < \beta < 0.9$ and $\gamma$ was standard normal.

After a new essential graph $G'$ was learned for an original graph $G$, four quantities were calculated: $Adj_+$ is the number of adjacencies present in $G'$ but not in $G$, $Adj_-$ is the number of adjacencies present in $G$ but not in $G'$, $Arcs_+$ is the number of directed arcs present in $G'$ but not in $G$, and $Arcs_-$ is the number of directed arcs present in $G$ but not in $G'$. Thus, if $G' = G$ $Adj_+ = Adj_- = Arcs_+ = Arcs_- = 0$.

**Experiment 1.** The first experiment compared the four algorithms {EGS, EGS/GS, GS, GS/1} by running each algorithm {50, 20, 10} times for datasets containing {15, 30, 45} variables, respectively. In each experiment the generating graph had $\overline{M}_{arcs} = 1.5$ times the number of nodes, and each database had 500 records. The results of this experiment are shown in Figure 4.

It is clear from this figure that EGS consistently outperforms all the other algorithms. Furthermore, EGS/GS consistently outperforms GS, but EGS/GS and GS/1 are often comparable. Note in each figure that the vertical axis is scaled to the maximum number of nodes. There appears to be little dependence on the number of nodes, although EGS may perform relatively better on graphs with many nodes.

The evident performance difference between EGS and EGS/GS seems to imply that search time is much better spent searching the space of essential graphs than searching the space of dags. A considerable majority of the search time (well over 90%) in the EGS/GS algorithm was spent in the greedy dag search.

**Experiment 2.** The second experiment investigated the dependence of the algorithm results on the number of records in the database. 50 dags were generated each with 15 nodes and $\overline{M}_{arcs} = 22$. Four databases were generated with $N$ records for $N = \{250, 500, 1000, 2000\}$. From these databases an essential graph was learned using each of the four algorithms {EGS, EGS/GS, GS, GS/1}. These results are shown in Figure 5 and Figure 6. Figure 5 shows the average results for the EGS algorithm only. Figure 6 shows the average value of the quantity $Error = Adj_+ + Adj_- + Arcs_+ + Arcs_-$ for each of the four algorithms as a function of the size of the data set. The EGS and EGS/GS algorithms show dramatic improvement as the number of records in-creases compared to GS and GS/1. It is evident from Figure 6 however, that none of the algorithms using a greedy search have converged. This fact is evident because the error term is not decreasing as a function of increasing data size.

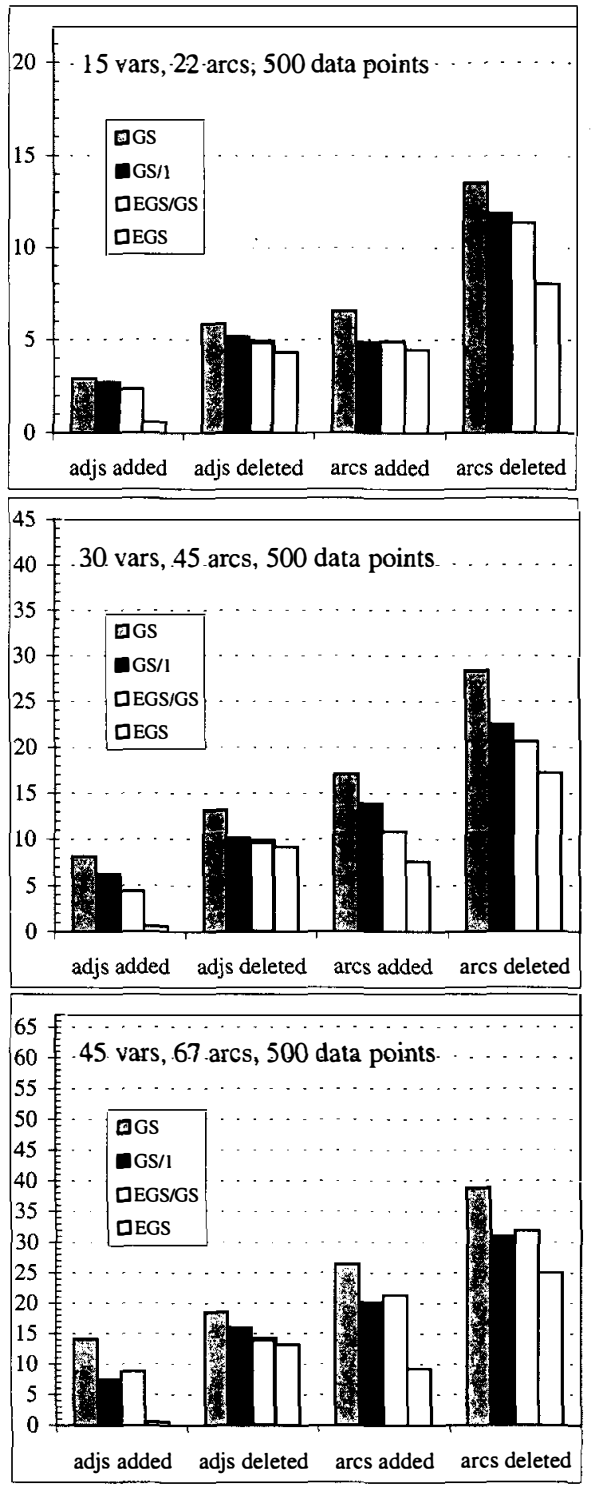

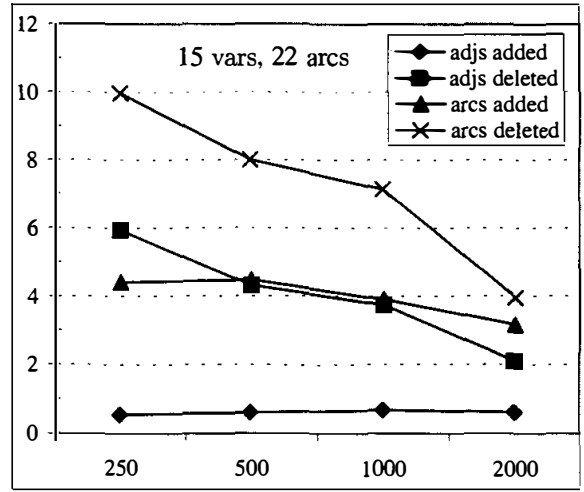

Figure 5: The dependence of $Adj_+$, $Adj_-$, $Arcs_+$, and $Arcs_-$ on the number of records in the database for EGS, for a study with generating graphs having 15 nodes and $\overline{M}_{arcs} = 22$.

Figure 4: Experimental results showing the mean number of adjacencies added, adjacencies deleted, arcs added and arcs deleted for each of the four algorithms. Each generating graph had 15, 30, and 45 nodes with $\overline{M}_{arcs}$ =22, 45, and 67, respectively.

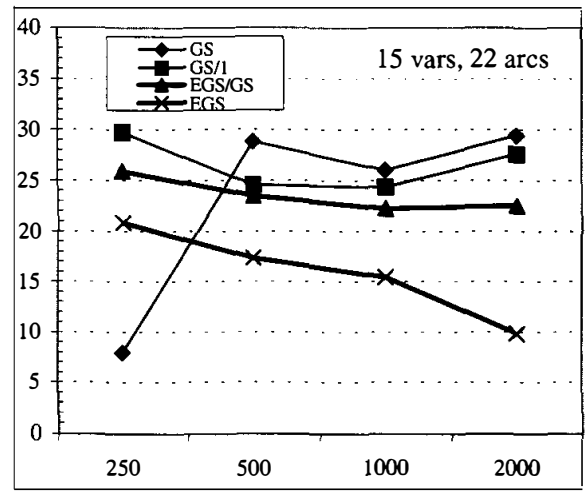

Figure 6: The average total error of all algorithms as a function of the number of records in the database.

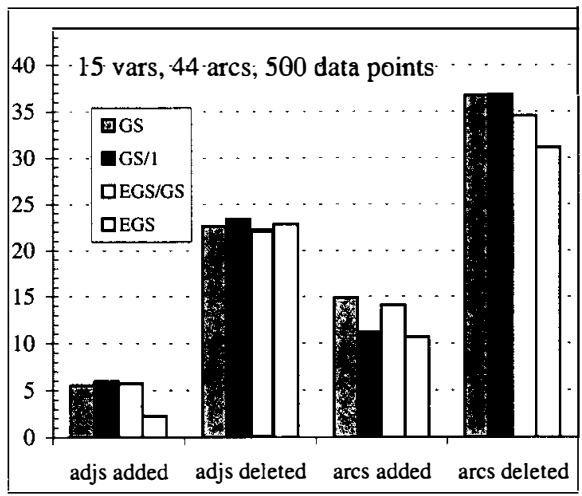

Figure 7: Experimental results showing the mean number of adjacencies added, adjacencies deleted, arcs added and arcs deleted for each of the four algorithms. Each generating graph had 15 nodes with $\overline{M}_{arcs} = 44$ and each data set had 500 records.

**Experiment 3.** The final experiment tested the variation of the quality of the algorithms with the density of the generating graph. The algorithms were tested using 50 generating graphs of 15 nodes and $\overline{M}_{arcs} = 44$. 500 records were generated for each test. The results of this experiment are shown in Figure 7. Comparing this graph to Figure 4a, it is evident that EGS and EGS/GS are not as effective on dense graphs. This is expected because of the added complexity incurred by constraint-based algorithms for dense graphs. Nonetheless, the graphs generated in this experiment were quite dense by common standards, over twice as dense as the ALARM network, a common benchmark model, and still EGS outperformed GS and GS/1.

## 5 Conclusions and Future Work

The idea behind this work is a simple one that combines already well-established ideas into yet another variant on search and score methods. Still it is interesting that the technique of searching over the arbitrary parameters in the otherwise deterministic constraint-based algorithms has not been tried previously. How powerful this method will turn out to be remains to be seen. Unfortunately, there are many possible criteria by which we could judge the efficacy of this algorithm, and whether one is much better than another is not clear. One common approach is to attempt to learn a gold standard network, such as the ALARM network. The value of this approach for the EGS algorithm is probably limited however, because EGS is only expected to perform well for small data sets, but most previous tests on the ALARM net have not focused on the sparse data regime.

We have, however, provided evidence that the EGS algorithm is worth investigating more closely. It consistently out performs the two general variants of greedy search algorithms considered in this paper. Furthermore, the marked improvement of EGS over the EGS/GS variant is even more evidence that the essential graph search is more efficient than the greedy search. Nonetheless there are many variations of these algorithms that could be compared to. It remains to see how EGS compares to these algorithms as well. Also, as mentioned in Section 3, there are many variations of EGS itself that could be tried.

EGS and EGS/GS assumed no latent variables. An obvious extension would be to include latent variables in the search. This could be done most effectively by using the FCI algorithm as the heuristic for the essential graph search. The resulting graphs could be scored with a Bayesian score using techniques for dealing with missing data, for example the EM algorithm using the BIC criterion, or using stochastic sampling methods. It may also be worthwhile to find a more principled approach to selecting the distribution from which the significance level is drawn.

Overall, techniques combining constraint-based and Bayesian approaches to learning, as well as approaches for searching over the space of essential graphs continue to show promise.

### Acknowledgments

We would like to thank the UAI reviewers for their insightful and incisive comments. This research was supported by the Air Force Office of Scientific Research under grant number F49620-97-1-0225 to University of Pittsburgh, and by the National Science Foundation under Faculty Early Career Development (CAREER) Program, grant IRI-9624629.